\documentclass[letterpaper,journal]{IEEEtran}

\usepackage{amsmath,amsfonts,amssymb}
\usepackage{array}
\usepackage{bm}
\usepackage{booktabs}
\usepackage{graphicx}
\usepackage{multirow}
\usepackage{stfloats}
\usepackage{tabularx}
\usepackage{textcomp}
\usepackage{url}
\usepackage{cite}
\usepackage[hidelinks]{hyperref}

\usepackage{amsmath,amsfonts,bm}

\def\eqref#1{equation~\ref{#1}}

\def\1{\bm{1}}

\DeclareMathAlphabet{\mathsfit}{\encodingdefault}{\sfdefault}{m}{sl}
\SetMathAlphabet{\mathsfit}{bold}{\encodingdefault}{\sfdefault}{bx}{n}

\newcolumntype{Y}{>{\raggedright\arraybackslash}X}
\newcolumntype{C}[1]{>{\centering\arraybackslash}p{#1}}

\newcommand{\papertablestyle}{\scriptsize\setlength{\tabcolsep}{3pt}\renewcommand{\arraystretch}{1.12}}
\newcommand{\papertablestyletight}{\scriptsize\setlength{\tabcolsep}{2pt}\renewcommand{\arraystretch}{1.12}}

\title{TurboT2VA: Fast Large-Scale Text-to-Video-Audio Generation via Score-Regularized Consistency Distillation}

\author{
Xiaoda Yang$^{*}$,
Yuxiang Liu$^{*}$,
Kaiwen Zheng,
Yuan Liu,
Yibo Lai,
Shengpeng Ji,
Kai Jiang,
Jianfei Chen,
Shan Yang,
Sen Liang,
Xiaobin Hu,
Shuicheng Yan,
Jintao Zhang$^{\dagger}$,
Jun Zhu$^{\dagger}$,
Zhou Zhao$^{\dagger}$
\thanks{$^{*}$Equal contribution.}
\thanks{$^{\dagger}$Corresponding authors.}

\thanks{Xiaoda Yang, Yibo Lai, Shengpeng Ji, and Zhou Zhao are with Zhejiang University, Hangzhou, China. E-mail: 1992426088@qq.com, 22621244@zju.edu.cn, shengpengji@zju.edu.cn, zhaozhou@zju.edu.cn}

\thanks{Yuxiang Liu is with Tianjin University, Tianjin, China. E-mail: lyx1021@tju.edu.cn}

\thanks{Yuan Liu is with Qingdao University, Qingdao, China. E-mail: \mbox{liuyuan2@qdu.edu.cn}}

\thanks{Kaiwen Zheng, Kai Jiang, Jianfei Chen, Jintao Zhang, and Jun Zhu are with Tsinghua University, Beijing, China. E-mail: zkwthu@gmail.com, ccku777@gmail.com, jianfeic@tsinghua.edu.cn, zhang-jt24@mails.tsinghua.edu.cn, dcszj@tsinghua.edu.cn}

\thanks{Shan Yang is with Tencent. E-mail: shaanyang@tencent.com}

\thanks{Sen Liang is with University of Science and Technology of China, Hefei, China. E-mail: liangsen@mail.ustc.edu.cn}

\thanks{Xiaobin Hu and Shuicheng Yan are with National University of Singapore, Singapore. E-mail: ben0xiaobin0hu1@nus.edu.sg, yansc@nus.edu.sg}

}

\begin{document}

\maketitle

\begin{figure*}[!t]
  \centering
  \includegraphics[width=\linewidth]{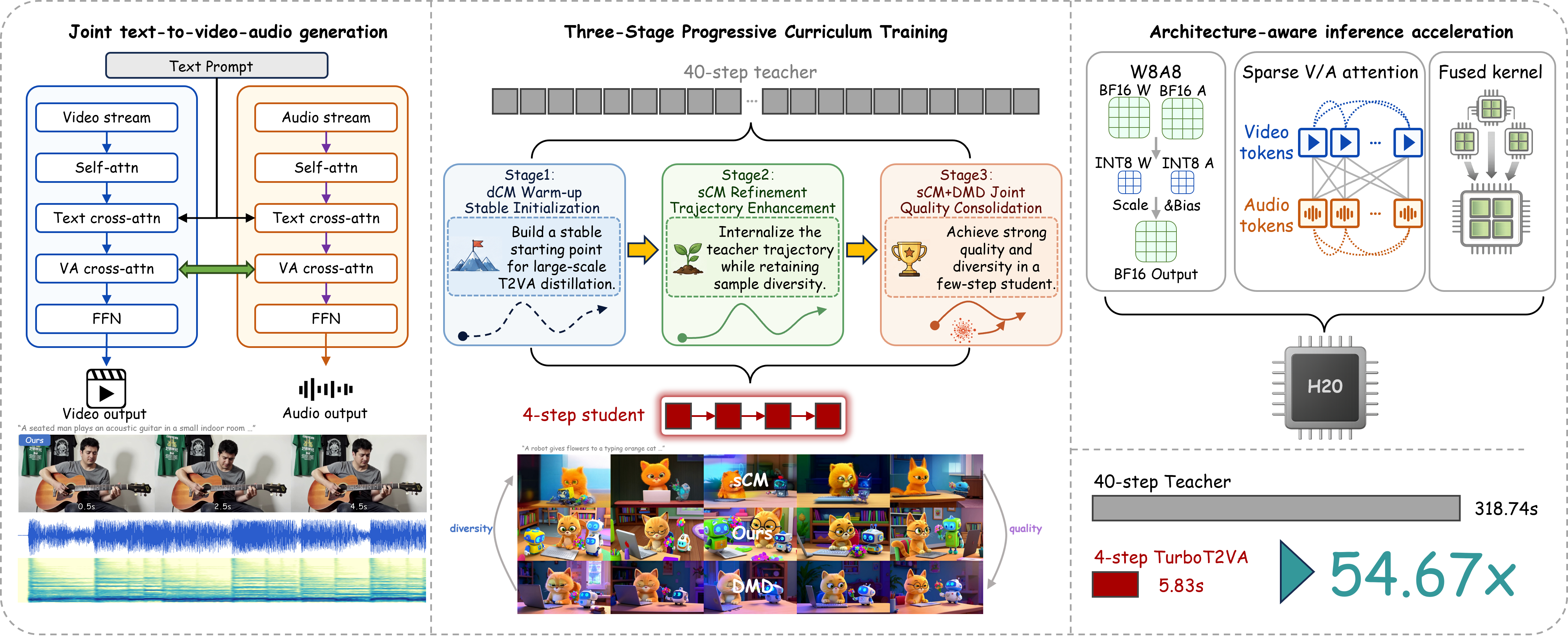}
  \caption{
  Overview of TurboT2VA. A joint video-audio Transformer is distilled from a 40-step teacher into a four-step student through a progressive dCM$\rightarrow$sCM$\rightarrow$sCM+DMD curriculum, improving the quality--diversity trade-off for synchronized video-audio generation. The distilled student is further accelerated by an architecture-aware inference stack consisting of modality-aware sparse attention dispatch, guarded W8A8 linear operators, and fused Transformer kernels, yielding a 54.67$\times$ generator-only speedup under the 1024$\times$1792 high-resolution deployment setting on one NVIDIA H20.
  }
  \label{fig:teaser}
\end{figure*}

\begin{abstract}
Joint text-to-video-audio generation produces synchronized visual and acoustic content, but the long sampling trajectories and heterogeneous multimodal computation of large models make inference prohibitively expensive. We present TurboT2VA, a distillation and inference framework for accelerating a 19B-parameter joint video-audio model. Compared with discrete-time consistency, continuous-time consistency avoids finite-step discretization errors in the training target and provides more faithful supervision along the teacher trajectory, making it particularly attractive for few-step distillation. However, scaling continuous-time consistency to a 19B joint T2VA model is challenged by modality-imbalanced optimization, numerically fragile trajectory-tangent estimation, and the quality--diversity trade-off. TurboT2VA addresses these issues with per-modality normalization and a progressive dCM$\rightarrow$sCM$\rightarrow$sCM+DMD curriculum, where dCM establishes a stable initialization, sCM more faithfully learns the continuous teacher trajectory while preserving diversity, and DMD improves perceptual quality. On LTX-2, four-step distillation reduces generator latency from 50.52s to 2.51s at the standard evaluation resolution of 512$\times$768, achieving a 20.1$\times$ speedup while maintaining strong visual quality, audio fidelity, diversity, and video-audio synchronization. We further develop an architecture-aware inference stack that combines guarded W8A8 and fused operators, padded-text compaction, and modality-aware sparse attention while preserving dense cross-modal and text-conditioning paths. Under the high-resolution deployment setting at 1024$\times$1792, the complete stack reduces generator latency from 318.74s to 5.83s on one NVIDIA H20, achieving a 54.67$\times$ generator-only speedup. Code and demos are available at \url{https://github.com/thu-ml/TurboDiffusion/tree/main/turbot2va}.
\end{abstract}



\section{Introduction}

\textbf{Motivation.}
Text-to-video-audio (T2VA) generation has recently made significant progress in producing visually realistic videos with synchronized audio, with models such as LTX-2 \cite{hacohen2026ltx} demonstrating the promise of unified audio-visual generation. However, these models rely on multi-step diffusion or flow-based sampling, resulting in high inference cost and limiting real-world deployment. To reduce this cost, existing methods mainly follow two directions. Consistency-based distillation enables few-step sampling by enforcing compatible predictions across noise levels. Within this family, dCM enforces consistency between predictions at finitely separated timesteps and typically relies on numerical PF-ODE updates over a prescribed timestep grid. This finite-step construction introduces discretization errors and requires careful scheduling of the timestep grid~\cite{song2023consistency,lu2025simplifying}. In contrast, sCM takes the continuous-time limit and directly follows the local tangent of the teacher trajectory, avoiding the finite-step target approximation and providing more faithful supervision; it has also been shown to outperform discrete-time consistency across different discretization settings~\cite{lu2025simplifying}. Such faithful continuous-trajectory supervision is particularly attractive for T2VA distillation, where visual motion, acoustic evolution, and their temporal correspondence must be preserved jointly. Distribution Matching Distillation (DMD), in contrast, aligns student and teacher distributions via score discrepancies ~\cite{yin2024one,yin2024improved}. The score-regularized continuous-time consistency model (rCM) further combines sCM with score distillation to improve the quality--diversity trade-off~\cite{zheng2026large}, yet its extension to large-scale joint T2VA generation remains largely underexplored.

\textbf{Challenge.}
Directly applying T2I/T2V acceleration methods to T2VA faces three challenges. (C1) \textbf{Modality imbalance.} Video and audio differ in scale and temporal resolution, and naive joint loss leads to video-dominated optimization, hurting audio quality and synchronization. (C2) \textbf{Stable continuous-time trajectory learning.} While sCM provides more faithful continuous-time supervision without the finite-step approximation of dCM, its trajectory-tangent estimation via JVPs becomes numerically fragile when scaled to a 19B joint model, especially for audio, where small trajectory errors can disrupt acoustic evolution and video-audio synchronization. (C3) \textbf{Quality--diversity trade-off.} Consistency improves diversity while distribution matching improves realism but reduces diversity; balancing both in T2VA remains unresolved.

\textbf{Approach.}
To address these challenges, we propose \textbf{TurboT2VA}, a fast generation framework for large-scale T2VA generation, as summarized in Figure~\ref{fig:teaser}. TurboT2VA extends score-regularized continuous-time consistency distillation to large-scale joint T2VA generation. (A1) To mitigate modality imbalance, we decouple video and audio losses and apply per-modality normalization, preventing dominant video gradients from overwhelming audio learning while preserving cross-modal interaction through shared generation and joint objectives. (A2) To retain the benefits of sCM while providing a more stable starting point, we adopt a progressive dCM$\rightarrow$sCM training strategy: dCM provides a stable initialization, after which sCM refines the student with more faithful continuous teacher-trajectory supervision. (A3) Based on this stabilized optimization process, we jointly optimize consistency and distribution matching, which balances trajectory preservation and perceptual quality enhancement. This design maintains the diversity benefits of consistency learning while leveraging distribution matching to improve realism, avoiding degradation caused by premature distribution-level supervision.

\textbf{Performance.}
We implement TurboT2VA based on LTX-2 and conduct large-scale distillation on a 19B-parameter T2VA model consisting of a 14B video backbone and a 5B audio backbone. At the standard evaluation resolution of 512$\times$768, four-step distillation reduces average inference time from 50.52s to 2.51s, achieving a 20.1$\times$ speedup while maintaining a strong overall trade-off across visual quality, audio fidelity, sample diversity, and video-audio synchronization. Under the high-resolution deployment setting at 1024$\times$1792, the complete architecture-aware inference stack reduces generator latency from 318.74s for the 40-step teacher to 5.83s for the accelerated student, yielding a 54.67$\times$ generator-only speedup on one H20. We evaluate TurboT2VA against a broad set of baselines, including cascaded generation pipelines, open-source and closed-source T2VA models, and ablations of our distillation strategy, demonstrating its effectiveness for practical large-scale T2VA acceleration, while remaining competitive across diverse evaluation settings and metrics.

\textbf{Contributions.}
Our contributions are as follows:
\begin{itemize}
\item We propose \textbf{TurboT2VA}, the first score-regularized consistency distillation framework for a 19B-parameter joint T2VA generation model, enabling efficient few-step video-audio generation.
\item We introduce joint cross-modal distillation, where video and audio are distilled through a shared process instead of independent branches, preserving temporal consistency.
\item We design a progressive curriculum distillation paradigm that stabilizes large-scale training and balances consistency-based diversity preservation with distribution-matching quality enhancement.
\item We develop an architecture-aware inference stack, combining guarded W8A8 and fused operators with modality-aware sparse attention to achieve a 54.67$\times$ generator-only speedup under the 1024$\times$1792 high-resolution deployment setting on one H20.
\end{itemize}

\section{Preliminary}

\subsection{From Video Generation to Joint Video-Audio Generation}

Recent text-to-video models have made substantial progress in visual fidelity, motion coherence, and temporal consistency
\cite{singer2022make,yang2025cogvideox,hacohen2024ltx,zheng2024open}.
However, most of them generate silent videos, whereas realistic multimedia content requires audio that is both semantically relevant and temporally aligned with visual events. Related audio-visual studies have also explored using audio data and generated-data augmentation to improve video speech recognition~\cite{yang2024audiovsr}. Cascaded pipelines partially address this limitation by generating audio from video
\cite{jeong2025read,zhang2026foleycrafter,cheng2025mmaudio}
or video from audio
\cite{yariv2024diverse,jeong2023power}.
Nevertheless, modeling the two modalities sequentially may introduce semantic mismatch, rhythmic inconsistency, and temporal misalignment.

Joint video-audio generation instead models visual and acoustic content in a unified process. Related audio-visual co-generation tasks have also explored cross-modal correspondence between music and visual dynamics~\cite{yang2026tmd}. Representative systems such as JavisDiT, OVI, LTX-2, and DaVinci-MagiHuman, synthesize video and audio using unified architectures or coupled modality-specific streams
\cite{liu2026javisdit,low2025ovi,hacohen2026ltx,chern2026speed}.
Although such designs enable explicit cross-modal interaction and improve video-audio coherence, they also incur substantially higher inference costs because both modalities must be generated through long diffusion or flow trajectories. Distilling a multi-step T2VA teacher into a few-step student is therefore important for practical deployment.

\subsection{Diffusion and Flow Distillation for Fast Generation}

Existing acceleration methods for diffusion and flow-based models largely
fall into two categories. Consistency-based distillation enables one- or
few-step generation by enforcing compatible predictions across noise levels.

Within this family, dCM enforces consistency between predictions at discretized noise levels, whereas sCM formulates consistency directly in continuous time. Because dCM constructs targets between finitely separated timesteps and relies on numerical PF-ODE transitions, its supervision is affected by timestep-grid design and finite-step discretization errors. sCM instead follows the infinitesimal tangent of the teacher trajectory, avoiding this finite-step approximation and providing more faithful training targets~\cite{song2023consistency,lu2025simplifying}. This benefit, however, comes with additional optimization difficulty: sCM requires trajectory-tangent estimation through Jacobian--vector products (JVPs), which makes large-scale training more numerically demanding~\cite{lu2025simplifying,zheng2026large}. Distribution-matching distillation instead aligns student and teacher distribution through score-level discrepancies.
Representative methods include DMD and its improved variant DMD2, which
achieve strong one- and few-step synthesis quality but may exhibit
mode-seeking behavior and reduce sample diversity
\cite{yin2024one,yin2024improved,zheng2026large}. This trade-off is particularly important for T2VA generation, where diversity spans visual appearance, motion trajectories, rhythm patterns, sound events, and video-audio correspondence.

These two families therefore provide complementary strengths: consistency learning preserves generation trajectories and sample diversity, while distribution matching improves perceptual quality through teacher-distribution alignment. TurboT2VA combines these to achieve fast, high-quality, and synchronized few-step video-audio generation.

\begin{figure*}[t]
    \centering
    \includegraphics[width=0.98\linewidth]{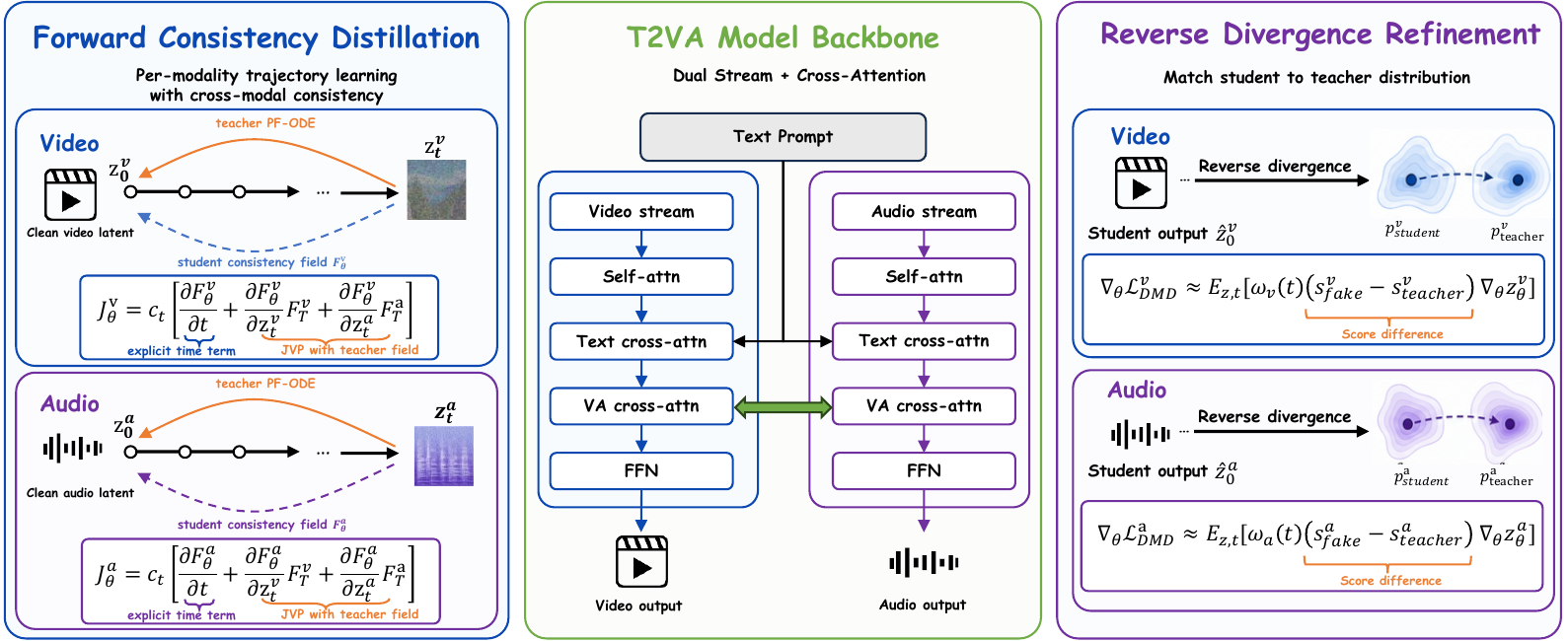}
    \caption{
    Overview of the proposed Dual-Divergence Distillation framework for T2VA acceleration. 
    The left panel illustrates forward consistency distillation for video and audio modalities, 
    the middle shows the T2VA backbone with cross-modal interaction, 
    and the right panel depicts reverse divergence (distribution matching) refinement.
    }
    \label{fig:dual_divergence}
\end{figure*}

\section{Cross-Modal Joint Distillation}

TurboT2VA adapts score-regularized consistency distillation from video-only generation to unified text-to-video-audio generation. Instead of distilling video and audio as two independent branches, we formulate T2VA distillation on paired video-audio latents. Let $\mathcal{M}=\{v,a\}$ denote the set of video and audio modalities. Given a text condition $c$, each clean sample is represented as $z_0=(z_0^v,z_0^a)$. We sample a shared base timestep $t$ and construct noisy latents by
\begin{equation}
z_t^m=\cos(t)z_0^m+\sin(t)\epsilon^m,\quad m\in\mathcal{M}.
\end{equation}
Here $\epsilon^v$ and $\epsilon^a$ are independently sampled noises, while the timestep and text condition are shared. Thus, video and audio are distilled along the same generation trajectory rather than two unrelated denoising paths.

The student predicts both modalities in one forward pass:
\begin{equation}
(\hat{z}_{0,\theta}^v,\hat{z}_{0,\theta}^a)=G_\theta(z_t^v,z_t^a,t,c).
\end{equation}
Although the losses are computed per modality, the forward graph is shared: the same T2VA Transformer receives paired latents, shared conditioning and timestep embeddings. Therefore, gradients from video and audio distillation losses are propagated through the same cross-modal Transformer, allowing alignment-relevant interactions to be optimized implicitly.

\subsection{Joint Continuous-Time Consistency}

For each modality $m\in\mathcal{M}$, the teacher and student clean predictions are converted into TrigFlow fields:
\begin{equation}
F_q^m=\frac{\cos(t)z_t^m-\hat{z}_{0,q}^{m}}{\sin(t)},\quad q\in\{T,\theta\}.
\end{equation}
Here $F_T^m$ defines the teacher trajectory, while $F_\theta^m$ is the student field to be optimized. sCM further requires a joint JVP of the student field along the paired teacher direction. With $c_t=\cos(t)\sin(t)$, we compute
\begin{equation}
\begin{aligned}
J_\theta^m
=
c_t\Bigg(
\frac{\partial F_\theta^m}{\partial t}
+
\frac{\partial F_\theta^m}{\partial z_t^v}F_T^v
+
\frac{\partial F_\theta^m}{\partial z_t^a}F_T^a
\Bigg),
\quad m\in\mathcal{M}.
\end{aligned}
\end{equation}
Since this JVP passes through the same cross-modal model, the video direction depends on audio tokens and the audio direction depends on video tokens. This is the key difference between joint T2VA distillation and simply adding two independent modality losses.

We define the sCM direction as

\begin{equation}
\begin{aligned}
g_{\mathrm{sCM}}^m
=&-\beta\cos(t)\sqrt{1-\gamma^2\sin^2(t)}
\left(\operatorname{sg}(F_\theta^m)-F_T^m\right)\\
&-\gamma\cos(t)\sin(t)z_t^m
-J_\theta^m ,
\end{aligned}
\end{equation}

where $\beta$ is the consistency boost factor, $\gamma$ is the tangent warm-up ratio, and $\operatorname{sg}(\cdot)$ denotes stop-gradient. For modality-balanced sCM, we normalize the consistency direction separately for each modality:
\begin{equation}
\bar{g}_{\mathrm{sCM}}^m
=
\frac{g_{\mathrm{sCM}}^m}{\left\|g_{\mathrm{sCM}}^m\right\|_2+\epsilon_{\mathrm{sCM}}},
\quad m\in\mathcal{M}.
\end{equation}
The norm is computed per sample over all latent elements of the corresponding modality. We define the sCM residual as
\begin{equation}
r_{\mathrm{sCM}}^m=
F_\theta^m-\operatorname{sg}(F_\theta^m)-\bar{g}_{\mathrm{sCM}}^m .
\end{equation}
The joint sCM loss is
\begin{equation}
\mathcal{L}_{\mathrm{sCM}}^{\mathrm{joint}}
=
\sum_{m\in\mathcal{M}}w_m\left\|r_{\mathrm{sCM}}^m\right\|_2^2 .
\end{equation}
This objective transfers the teacher's paired video-audio trajectory to the few-step student, preserving visual motion, acoustic evolution, rhythm structure, and temporal correspondence.

\subsection{Joint Distribution Matching}

sCM preserves the teacher trajectory, but few-step generation may still lose perceptual quality. We therefore introduce DMD after the student has learned a stable joint trajectory. Given an on-policy paired sample generated by the student,
\begin{equation}
\hat{z}_0=(\hat{z}_0^v,\hat{z}_0^a),
\end{equation}
we perturb it to $\hat{z}_t$ and compute the prediction discrepancy for each modality:
\begin{equation}
g_{\mathrm{DMD}}^m
=
\hat{z}_{0,\mathrm{fake}}^m(\hat{z}_t,t,c)
-
\hat{z}_{0,T}^m(\hat{z}_t,t,c),
\quad m\in\mathcal{M}.
\end{equation}
Here $\hat{z}_{0,T}$ denotes the teacher prediction, and $\hat{z}_{0,\mathrm{fake}}$ is predicted by the fake-score network trained on student-generated samples rather than by the generator itself. Under this model parameterization, this prediction discrepancy is equivalent to a score discrepancy. In practice, the teacher prediction is computed with classifier-free guidance, and video and audio can use different guidance scales.

Unlike sCM, DMD uses a residual-scale normalizer rather than an L2 normalizer. For each modality, we compute
\begin{equation}
c_{\mathrm{DMD}}^m
=
\frac{1}{N_m}\sum_{i\in m}
\left|
\hat{z}_{0,i}^m-\hat{z}_{0,T,i}^m
\right|,
\end{equation}
where $N_m$ is the number of latent elements in modality $m$. The normalized DMD gradient is
\begin{equation}
\bar{g}_{\mathrm{DMD}}^m
=
\frac{g_{\mathrm{DMD}}^m}{c_{\mathrm{DMD}}^m+\epsilon_{\mathrm{DMD}}},
\quad m\in\mathcal{M}.
\end{equation}
We define the DMD residual as
\begin{equation}
\Delta_{\mathrm{DMD}}^m
=
\hat{z}_0^m-
\operatorname{sg}
\left(
\hat{z}_0^m-\bar{g}_{\mathrm{DMD}}^m
\right).
\end{equation}

We aggregate the DMD surrogate with a sum reduction over latent elements:
\begin{equation}
\mathcal{L}_{\mathrm{DMD}}^{\mathrm{joint}}
=
\sum_{m\in\mathcal{M}}
w_m
\sum_{i\in m}
\left(\Delta_{\mathrm{DMD},i}^{m}\right)^2 .
\end{equation}
This preserves the accumulated distribution-matching signal over the full paired video-audio latent, allowing DMD to provide a sufficiently strong teacher-distribution anchor during joint sCM+DMD optimization. The overall scale is still controlled by modality-wise gradient normalization, modality weights, and the outer coefficient $\lambda_{\mathrm{DMD}}$.

Unlike applying DMD to video and audio independently, this term is computed from paired video-audio samples generated by the same student under the same condition and timestep. Therefore, DMD improves final-sample realism without breaking the cross-modal structure learned by joint sCM.

\subsection{Modality-Balanced Joint Optimization}

A direct joint objective can be dominated by one modality because video and audio latents have different dimensionalities, temporal resolutions, and learning speeds. TurboT2VA therefore applies modality-wise balancing to both objectives, but with different normalizers: sCM uses an L2 direction normalizer, while DMD uses a teacher-residual scale normalizer. The normalized modality losses are then combined with weights $w_m$. This balancing is applied only at the loss level; the student is still trained as a single unified T2VA generator. When sCM and DMD are optimized together, we reuse the same batch, text condition, and paired video-audio latents for both objectives, making the two objectives operate on the same paired samples and conditions.

The final joint objective is
\begin{equation}
\mathcal{L}_{\mathrm{stage3}}
=
\lambda_{\mathrm{sCM}}\mathcal{L}_{\mathrm{sCM}}^{\mathrm{joint}}
+
\lambda_{\mathrm{DMD}}\mathcal{L}_{\mathrm{DMD}}^{\mathrm{joint}} .
\end{equation}
The sCM term preserves the paired teacher trajectory, supporting diversity and video-audio synchronization under few-step sampling, while the DMD term improves final-sample quality by matching the teacher distribution. The outer coefficients $\lambda_{\mathrm{sCM}}$ and $\lambda_{\mathrm{DMD}}$ further control the balance between trajectory consistency and distribution-level anchoring.

\subsection{LTX-2 TrigFlow Adaptation}

The original LTX-2 backbone follows a rectified-flow velocity
parameterization, while sCM uses the TrigFlow parameterization. We introduce a wrapper that maps TrigFlow latents and timesteps to the LTX-2 rectified-flow interface:
\begin{equation}
z_{\mathrm{rf}}=\frac{z_{\mathrm{trig}}}{\cos(t)+\sin(t)},\quad
t_{\mathrm{rf}}=\frac{\sin(t)}{\cos(t)+\sin(t)} .
\end{equation}
This wrapper exposes video and audio predictions under a unified timestep interface and supports JVP computation through timestep embeddings, video-audio tokenization, positional embeddings, and the cross-modal Transformer. We do not modify the underlying LTX-2 backbone; instead, TurboT2VA adapts its training interface so that joint sCM and DMD objectives can be applied to both modalities within one student model.

\section{Curriculum Distillation Paradigm}

\begin{figure}[t]
    \centering
    \includegraphics[width=\linewidth]{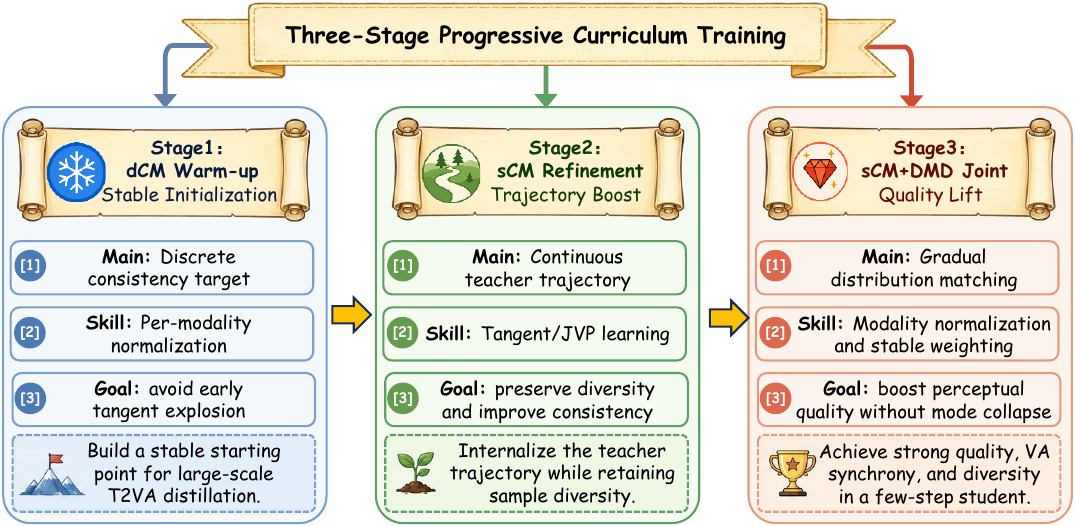}
    \caption{
    Overview of the proposed curriculum distillation paradigm.
    The student is progressively trained through dCM warm-up, sCM refinement, and sCM+DMD joint optimization.
    }
    \label{fig:curriculum}
\end{figure}

Although direct sCM+DMD training is feasible, we find that the ordering of trajectory learning and distribution matching substantially affects the final quality--diversity trade-off. We therefore investigate the optimization path and motivate a staged curriculum for large-scale T2VA distillation.

\subsection{Empirical Observation: Trainable but Suboptimal}

We first compare alternative optimization paths before prescribing a curriculum. Training sCM+DMD directly from the base initialization is feasible and reaches reasonable quality, showing that joint optimization itself is not unstable. However, as shown in Figure~\ref{fig:combined_ablation}(a), the staged route overtakes direct joint training after entering the final stage and maintains a Javis-score advantage at the common 7,500-step endpoint. Complementary objective ablations further show that sCM alone better preserves trajectory diversity and temporal structure but produces weaker final-sample quality, whereas DMD emphasizes perceptual quality at the cost of diversity. Introducing joint sCM+DMD only after dCM warm-up and sCM refinement yields the strongest overall balance among visual quality, audio fidelity, sample diversity, and video-audio synchronization.

These observations indicate that the benefit of curriculum distillation lies in structuring the optimization path. The student first learns how to follow the teacher trajectory, and only later receives stronger distribution-level supervision.

\subsection{Why Direct sCM+DMD is Suboptimal}

The empirical gap can be explained by the premature coupling of two objectives that play different roles. sCM transfers the teacher's generation trajectory to the student and encourages consistency across noise levels. This helps preserve temporal dynamics, motion evolution, audio events, and cross-modal correspondence. In contrast, DMD provides a distribution-level anchor that improves final-sample realism by matching the teacher distribution.

When DMD is introduced too early, distribution-level refinement may encourage the student to match teacher-like samples before it has learned a sufficiently broad trajectory prior. This does not necessarily cause training failure, but it can weaken the diversity inherited from the teacher trajectory. The effect is particularly important for T2VA generation, where diversity is expressed not only through visual appearance, but also through motion trajectories, rhythm patterns, sound events, and video-audio correspondence. The staged curriculum therefore first establishes a diverse and stable trajectory prior and only then introduces DMD to improve perceptual quality.

\subsection{Three-Stage Curriculum Distillation}

We propose a progressive curriculum that decomposes T2VA distillation into three stages. Similar progressive-training ideas have also been explored in vision-language reasoning~\cite{yang2026progressive}. Starting from the base student initialization $\theta_0$, the student parameters are updated in a staged manner:
\begin{equation}
\theta_1
=
\operatorname{Train}
\left(
\theta_0,\mathcal{L}_{\mathrm{dCM}}
\right),
\end{equation}
\begin{equation}
\theta_2
=
\operatorname{Train}
\left(
\theta_1,\mathcal{L}_{\mathrm{sCM}}
\right),
\end{equation}
\begin{equation}
\theta_3
=
\operatorname{Train}
\left(
\theta_2,
\lambda_{\mathrm{sCM}}\mathcal{L}_{\mathrm{sCM}}
+
\lambda_{\mathrm{DMD}}\mathcal{L}_{\mathrm{DMD}}
\right).
\end{equation}
Here $\operatorname{Train}(\theta,\mathcal{L})$ denotes continuing optimization from parameters $\theta$ under objective $\mathcal{L}$. This formulation emphasizes that the curriculum changes the optimization path, rather than the final model architecture.

\subsubsection{Stage 1: dCM Warm-up}

We first optimize a dCM objective over discrete noise levels to initialize the student for coarse denoising. Since dCM does not require continuous-time trajectory-tangent estimation, it provides a considerably more stable starting point for large-scale joint video-audio distillation. However, its finite-step consistency targets remain subject to discretization errors and provide only an approximation to the continuous teacher trajectory. We therefore use dCM as an optimization scaffold rather than the final consistency objective. This warm-up equips the student with basic denoising capability, stable timestep conditioning, and a coarse joint denoising prior for subsequent continuous trajectory learning.

\subsubsection{Stage 2: sCM Refinement}

Starting from the dCM-initialized student, we switch to sCM to learn the continuous teacher-guided generation trajectory. By taking the continuous-time limit, sCM avoids the finite-step target approximation of dCM and transfers the teacher's local trajectory directions more faithfully. This stage therefore upgrades the student from coarse discrete consistency to continuous trajectory learning. This refinement improves trajectory fidelity and preserves sample diversity before distribution-level supervision is introduced. For T2VA generation, such continuous trajectory supervision is particularly important because it affects visual motion, acoustic evolution, and video-audio temporal correspondence simultaneously.

\subsubsection{Stage 3: sCM+DMD Joint Optimization}

Inspired by the score-regularization principle of rCM
\cite{zheng2026large}, we introduce DMD for distribution refinement
while continuing sCM training after the student acquires sufficient
trajectory coverage. At this stage, sCM maintains trajectory structure and diversity, whereas DMD improves final-sample realism by pulling the student distribution closer to the teacher distribution. This staged introduction avoids applying strong distribution-level supervision before the student has learned a sufficiently expressive trajectory prior.

Overall, the dCM$\rightarrow$sCM$\rightarrow$sCM+DMD sequence converts direct joint optimization into a structured process that builds a coarse denoising prior, expands it into a trajectory-consistent generator, and finally improves distribution-level quality, yielding a stronger quality--diversity--synchronization trade-off.

\section{Architecture-Aware Acceleration for Joint Video-Audio Generation}
\label{sec:inference_acceleration}

Few-step distillation reduces the number of Transformer evaluations but not the cost of each evaluation. This per-step cost becomes dominant at high resolution, where the joint video-audio Transformer contains long modality-specific token sequences, heterogeneous attention paths, and large feed-forward projections. Directly applying a video-only inference stack is unsafe because the heterogeneous attention paths differ in sequence lengths, masks, and matrix shapes. Building on the efficient primitives of TurboDiffusion~\cite{zhang2025turbodiffusion}, we specialize an architecture-aware inference stack for LTX-2 comprising modality-aware SageSLA dispatch, shape-aware post-scale W8A8 linear operators, and fused multimodal Transformer operations. The stack is applied after distillation and requires no retraining of the TurboT2VA student.

\subsection{Modality-Aware Sparse Attention Dispatch}

The joint Transformer contains four semantically different attention paths: video self-attention, audio self-attention, bidirectional video-audio interaction, and masked text cross-attention. Applying the same approximation to all paths can alter conditioning behavior or violate sparse-kernel assumptions. Our dispatcher identifies modules by their architectural role and applies SageSLA~\cite{zhang2025sageattention,zhang2026sla} only to the unmasked video and audio self-attention paths. Bidirectional cross-modal attention and masked text cross-attention remain dense, preserving explicit video-audio exchange and text conditioning.

For each selected path, the adapter transforms the native sequence-major query, key, and value tensors into the layout required by SageSLA. A query-dependent block map retains a fraction $\rho_\ell$ of key blocks at layer $\ell$, with support for either one global ratio or a layer-wise schedule. On the H20 path, query and key are quantized to INT8 and value aggregation uses FP8 values. To avoid degenerate behavior on short audio sequences, the runtime ratio is
\begin{equation}
\tilde{\rho}_\ell=\max\left(\rho_\ell,\frac{1}{N_{\mathrm{blk}}}\right),
\end{equation}
where $N_{\mathrm{blk}}$ is the number of key blocks, guaranteeing at least one retained key block. The result is then restored to the original joint-Transformer layout. Our high-resolution configuration uses $\rho_\ell=0.3$ for all layers. This setting concentrates approximation on the long self-attention sequences that dominate computation without assuming equal video and audio sequence lengths.

\subsection{Shape-Aware Post-Scale W8A8 Linear Operators}

Feed-forward networks and attention projections remain a major cost after attention sparsification. Their matrix shapes differ across video, audio, and cross-modal branches, while a blockwise quantization strategy repeatedly applies scales along the reduction dimension. We instead use a post-scale W8A8 operator with static per-output-channel weight quantization and dynamic per-row activation quantization. For activation row $x_i$ and output-channel weight row $w_j$, we compute
\begin{align}
\hat{x}_i &= \operatorname{clip}\left(\operatorname{round}(x_i/s_i^x),-127,127\right), \\
\hat{w}_j &= \operatorname{clip}\left(\operatorname{round}(w_j/s_j^w),-127,127\right), \\
y_{ij} &= s_i^x s_j^w \sum_k \hat{x}_{ik}\hat{w}_{jk}+b_j ,
\end{align}
where $s_i^x=\max(\lVert x_i\rVert_\infty,\epsilon)/127$, $s_j^w=\max(\lVert w_j\rVert_\infty,\epsilon)/127$, and $\epsilon=10^{-4}$. The TileLang kernel accumulates INT8 products in INT32  across the full reduction dimension $K$, then applies both scales and bias once in the epilogue. This avoids fragmenting the reduction with tile-local rescaling.

The current optimized path requires CUDA BF16 inputs and flattened matrix dimensions $M$, $K$, and $N$ divisible by 128. Unsupported dtypes or shapes fall back to native BF16 linear computation using a retained BF16 weight copy. We additionally concatenate query, key, and value projections when they share the same input, and concatenate key/value projections in cross-attention, reducing redundant activation quantization and kernel launches. This shape-aware policy prioritizes safe coverage rather than forcing every heterogeneous branch through one quantized kernel.

\subsection{Fused Multimodal Transformer Operations}

After accelerating attention and linear operators, memory-bound elementwise operations form a non-negligible fraction of each denoising step. The joint Transformer repeatedly combines normalization with modality- and timestep-dependent modulation, followed by gated residual updates. We replace module-level RMSNorm and LayerNorm and fuse recurring functional patterns, including modulated RMS normalization, adaptive scale--shift modulation, gated residual updates, output modulation, and split rotary embeddings. Every fused kernel has dtype, contiguity, shape, and mask guards; unsupported layouts execute the original implementation.

We also eliminate redundant computation in the text-conditioning path. During batch-size-one inference, the shared text mask identifies valid prompt tokens. We compact both video- and audio-conditioning embeddings to this same valid-token set and then remove the padding mask before cross-attention. Unlike semantic prompt truncation, this transformation removes only padded positions and leaves every valid token unchanged. For larger batches the optimization is disabled, because different prompts need not share one valid-token pattern.

\paragraph{Default inference configuration}
For high-resolution batch-one inference, we apply SageSLA to video and audio self-attention with $\rho_\ell=0.3$, quantize all compatible generator linear layers with the TileLang post-scale backend, enable fused normalization and modulation kernels, and compact padded text context. At 1024$\times$1792 with 121 frames, the video latent has shape $[1,16,128,32,56]$, corresponding to 28,672 video self-attention tokens. Sparse attention is approximate, so the retention ratio should be revalidated when the resolution or prompt distribution changes. We use this configuration for the subsequent high-resolution latency evaluation.

\section{Experiments}

\begin{table*}[t]
\centering
\caption{
JavisBench comparison with representative T2VA systems under the standard-resolution evaluation setting. Best and second-best open-source results are \textbf{bold} and \underline{underlined}.
}
\label{tab:full_main}

\papertablestyletight
\resizebox{\linewidth}{!}{
\begin{tabular}{lcc|ccc|cccc|ccc|cc}
\toprule

\multirow{2}{*}{\textbf{Model}} &
\multirow{2}{*}{\textbf{Size}} &
\multirow{2}{*}{\textbf{Time (s)} $\downarrow$} &
\multicolumn{3}{c|}{\textbf{Perceptual Quality}} &
\multicolumn{4}{c|}{\textbf{Text Semantic Consistency}} &
\multicolumn{3}{c|}{\textbf{V-A Semantic Consistency}} &
\multicolumn{2}{c}{\textbf{V-A Alignment}} \\

\cmidrule(lr){4-6}
\cmidrule(lr){7-10}
\cmidrule(lr){11-13}
\cmidrule(lr){14-15}

& &
& \textbf{Visual} $\uparrow$
& \textbf{Motion} $\uparrow$
& \textbf{Audio} $\uparrow$

& \textbf{IB-TV} $\uparrow$
& \textbf{IB-TA} $\uparrow$
& \textbf{CLIP} $\uparrow$
& \textbf{CLAP} $\uparrow$

& \textbf{IB-AV} $\uparrow$
& \textbf{CAVP} $\uparrow$
& \textbf{AVH} $\uparrow$

& \textbf{Javis} $\uparrow$
& \textbf{Desync} $\downarrow$ \\

\midrule

\multicolumn{15}{c}{\rule[-0.4ex]{0pt}{2.4ex}\textit{Closed-source T2VA Models}} \\[-0.2ex]

Kling v3~\cite{kuaishou2026kling3}
& -- & --
& 3.478 & 1.104 & 5.499
& 0.246 & 0.154 & 0.309 & 0.421
& 0.231 & 0.793 & 0.210
& 0.177 & 0.907 \\

Sora 2~\cite{openai2025sora2}
& -- & --
& 2.836 & 0.185 & 4.802
& 0.252 & 0.161 & 0.308 & 0.434
& 0.240 & 0.792 & 0.217
& 0.187 & 0.414 \\

Veo 3~\cite{deepmind2026veo3}
& -- & --
& 4.221 & 0.735 & 5.534
& 0.219 & 0.178 & 0.310 & 0.469
& 0.351 & 0.798 & 0.326
& 0.285 & 0.336 \\

\midrule

\multicolumn{15}{c}{\rule[-0.4ex]{0pt}{2.4ex}\textit{Cascaded T2A + A2V}} \\[-0.2ex]

TempoToken~\cite{yariv2024diverse}
& 1.3B & 7.76
& 0.885 & -0.671 & --
& \textbf{0.298} & -- & 0.292 & --
& 0.133 & 0.796 & 0.111
& 0.071 & 1.021 \\

TPoS~\cite{jeong2023power}
& 1.0B & 18.01
& 1.592 & -0.330 & --
& 0.240 & -- & 0.261 & --
& 0.145 & 0.793 & 0.138
& 0.098 & 1.112 \\

\midrule

\multicolumn{15}{c}{\rule[-0.4ex]{0pt}{2.4ex}\textit{Cascaded T2V + V2A}} \\[-0.2ex]

ReWaS~\cite{jeong2025read}
& 0.6B & 40.49
& -- & -- & 4.124
& -- & 0.075 & -- & 0.309
& 0.113 & 0.791 & 0.107
& 0.083 & 0.772 \\

FoleyCrafter~\cite{zhang2026foleycrafter}
& 1.2B & 36.74
& -- & -- & 4.225
& -- & 0.079 & -- & 0.212
& 0.182 & 0.798 & 0.168
& 0.134 & 0.934 \\

MMAudio~\cite{cheng2025mmaudio}
& 0.1B & 15.16
& -- & -- & 4.308
& -- & \textbf{0.164} & -- & 0.401
& 0.190 & 0.794 & 0.180
& 0.142 & \underline{0.524} \\

\midrule

\multicolumn{15}{c}{\rule[-0.4ex]{0pt}{2.4ex}\textit{Open-source T2VA Models}} \\[-0.2ex]

JavisDiT~\cite{liu2026javisdit}
& 3.1B & 49.73
& 1.488 & 0.591 & 4.405
& 0.250 & \underline{0.145} & \underline{0.297} & \textbf{0.413}
& \underline{0.212} & 0.798 & 0.183
& 0.152 & 0.848 \\

OVI~\cite{low2025ovi}
& 11B & 76.79
& \underline{2.145} & 0.294 & \underline{5.024}
& 0.244 & 0.132 & \textbf{0.307} & \underline{0.407}
& 0.209 & 0.795 & \underline{0.206}
& \underline{0.175} & 0.638 \\

DaVinci-MagiHuman~\cite{chern2026speed}
& 15B & \underline{6.70}
& 0.854 & 0.382 & \textbf{5.245}
& 0.227 & 0.124 & 0.284 & 0.362
& 0.201 & \underline{0.800} & 0.181
& 0.145 & 0.730 \\

\midrule

\multicolumn{15}{c}{\rule[-0.4ex]{0pt}{2.4ex}\textit{Teacher and TurboT2VA}} \\[-0.2ex]

Teacher (40 steps)~\cite{hacohen2026ltx}
& 19B & 50.52
& 1.968 & \underline{0.783} & 4.902
& \underline{0.268} & 0.139 & 0.291 & 0.340
& 0.199 & 0.796 & 0.195
& 0.165 & 0.617 \\

\textbf{Ours (4 steps)}
& 19B & \textbf{2.51}
& \textbf{2.389} & \textbf{0.849} & 4.486
& 0.218 & \underline{0.145} & 0.292 & 0.347
& \textbf{0.229} & \textbf{0.803} & \textbf{0.224}
& \textbf{0.196} & \textbf{0.388} \\

\bottomrule
\end{tabular}
}
\end{table*}

\subsection{Experimental Setup}

We conduct experiments on a large-scale text-to-video-audio generation model built upon LTX-2. The teacher consists of a 14B video backbone and a 5B audio backbone, forming a 19B-parameter joint video-audio generation system. Our goal is to distill the original multi-step teacher into a few-step student while preserving visual quality, audio fidelity, temporal coherence, sample diversity, and video-audio synchronization.

\paragraph{Training configuration}
TurboT2VA is trained on 8 H20 GPUs using a training set of 100K text-video-audio samples at 512$\times$768 resolution with 121 frames. We use per-GPU batch size 1, resulting in an effective batch size of 8. The main 4-step student follows a three-stage curriculum: 2K dCM warm-up steps, 0.5K sCM refinement steps, and 4.5K joint sCM+DMD optimization steps. The selected student is therefore trained for 7K optimization steps in total. The main joint stage uses AdamW with a learning rate of $2\!\times\!10^{-5}$, $\beta_1=0.0$, $\beta_2=0.999$, weight decay 0.01, and equal sCM and DMD weights. Teacher predictions use video and audio guidance scales of 3.0 and 5.0, respectively. Training uses BF16 mixed precision, gradient checkpointing, and FSDP, and takes approximately 21 hours on 8 H20 GPUs, excluding evaluation and checkpoint sweep overhead.

\paragraph{Standard-resolution evaluation protocol and metrics}
For the main student comparisons, ablations, training curves, and sampling-step analysis, all variants are evaluated on the same 200-prompt split at 512$\times$768 resolution with 121 frames. For diversity evaluation, we use the same 200 prompts and generate 8 samples per prompt with different random seeds. Runtime is measured as the average pure inference time per generated sample, excluding model loading and evaluation overhead. We report JavisBench metrics~\cite{liu2026javisdit} covering perceptual quality, text semantic consistency, video-audio semantic consistency, and video-audio alignment. We further use VBench for video fidelity \cite{huang2024vbench}, TTA-Bench for audio quality \cite{wang2026tta}, MS-CLAP for text-audio alignment \cite{elizalde2023clap}, and within-prompt ImageBind cosine distances for diversity \cite{girdhar2023imagebind}.

\paragraph{High-resolution deployment protocol}
We separately profile the architecture-aware inference stack on one NVIDIA H20 at batch size one, 1024$\times$1792 output resolution, and 121 frames. Unless otherwise specified, latency refers to generator-only time and excludes checkpoint loading, decoding, muxing video and audio, and disk I/O. Table~\ref{tab:highres_component_latency} further reports component-wise generator latency. All stages in the systems comparison are cumulative. The two resolution profiles serve complementary purposes: the standard-resolution protocol enables comprehensive model-level comparisons and ablations, whereas the high-resolution protocol serves as a more demanding deployment setting that better exposes the computational bottlenecks of the large-scale Transformer and the benefits of architecture-aware acceleration. All comparisons are conducted within the same resolution profile.

\begin{figure*}[t]
    \centering
    \includegraphics[width=\linewidth]{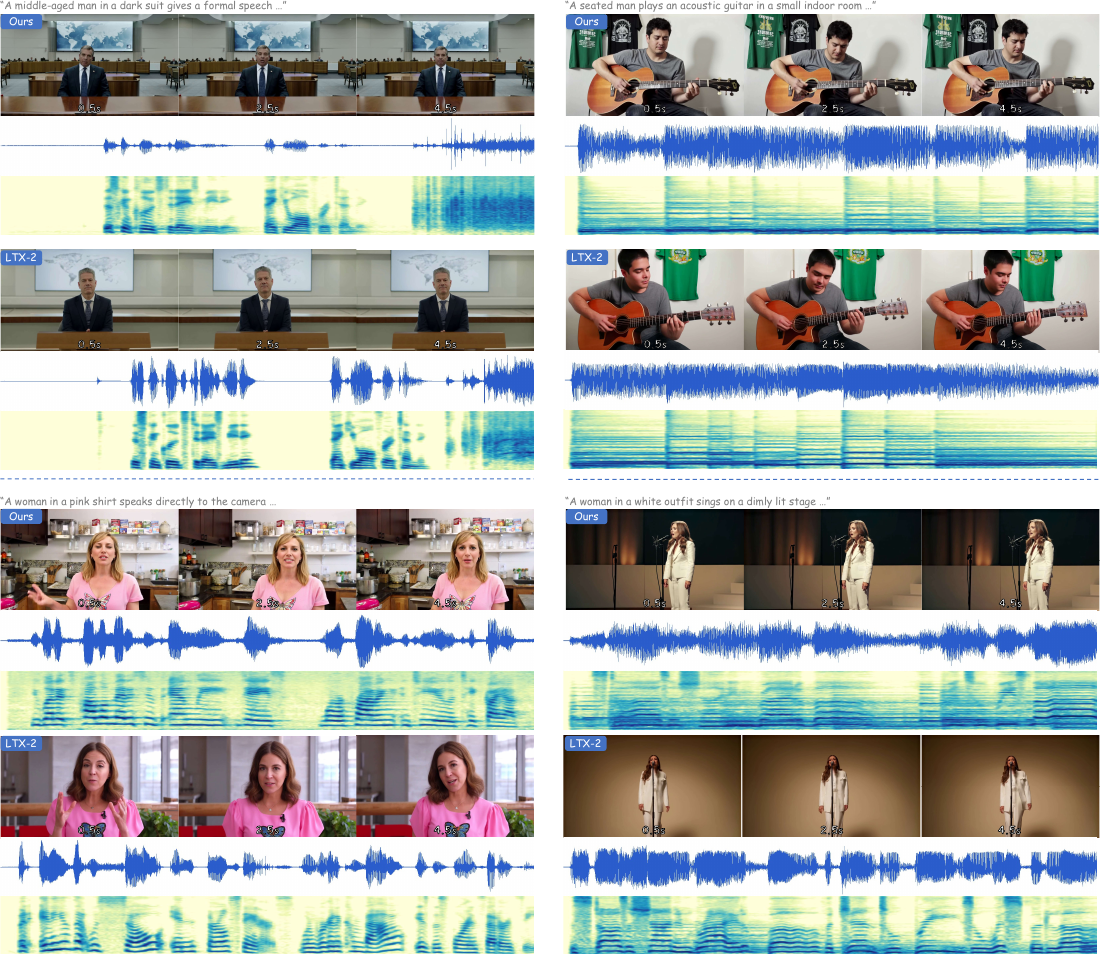}
    \caption{
    Qualitative video-audio examples. Each block compares the 40-step teacher and the 4-step TurboT2VA student under the same prompt, showing early, middle, and late video frames together with the corresponding waveform and mel spectrogram. The student preserves coherent subjects, scene layout, temporal appearance, and audio structure while reducing the sampling trajectory to four steps.
    }
    \label{fig:demo_display}
\end{figure*}

\subsection{Main Results}

\subsubsection{Generation Quality}
Table~\ref{tab:full_main} reports the main quantitative comparison on JavisBench and related video-audio metrics. TurboT2VA achieves a strong balance between inference efficiency and generation quality. Compared with the original 40-step teacher, TurboT2VA distills the generation process into a 4-step student, reducing the average inference time from 50.52s to 2.51s at the standard evaluation resolution of 512$\times$768 and achieving a 20.1$\times$ speedup. Despite the aggressive step reduction, TurboT2VA maintains comparable overall generation quality on JavisBench and achieves improved performance on several video-audio quality and synchronization metrics. Compared with cascaded, open-source, and closed-source T2VA baselines, TurboT2VA provides competitive generation quality while being substantially faster than most multi-step systems. These results demonstrate that the proposed curriculum distillation framework effectively preserves the teacher's generation capability and enables efficient few-step T2VA inference.

Figure~\ref{fig:demo_display} complements the quantitative results with qualitative temporal and audio comparisons. Across examples, the four-step student preserves the teacher's visual content and scene evolution while maintaining coherent audio structure.

\subsubsection{Video and Audio Fidelity}

We complement the joint evaluation with separate modality-specific fidelity measurements.

\paragraph{Video fidelity}
We evaluate video fidelity using VBench. Table~\ref{tab:vbench} reports the available VBench dimensions for the original teacher and ours, including aesthetic quality, imaging quality, motion smoothness, subject consistency, and temporal flickering. The distilled student maintains comparable video fidelity to the teacher on this auxiliary video-only benchmark, with improvements in aesthetic and imaging quality while preserving strong temporal consistency.

\begin{table}[t]
\centering
\caption{
VBench video fidelity evaluation between the original teacher and ours. The best result is \underline{underlined}.
}
\label{tab:vbench}
\papertablestyle
\begin{tabular*}{\linewidth}{@{\extracolsep{\fill}}lccccc@{}}
\toprule
\textbf{Model} &
\textbf{Aes.} $\uparrow$ &
\textbf{Img.} $\uparrow$ &
\textbf{Mot.} $\uparrow$ &
\textbf{Subj.} $\uparrow$ &
\textbf{Temp.} $\uparrow$ \\
\midrule
Teacher (40 steps) & 0.5286 & 0.5884 & \underline{0.9919} & 0.9592 & \underline{0.9844} \\
Ours (4 steps) & \underline{0.5519} & \underline{0.6520} & 0.9911 & \underline{0.9614} & 0.9773 \\
\bottomrule
\end{tabular*}
\end{table}

\paragraph{Audio fidelity}
We measure audio quality using TTA-Bench metrics, including content enjoyment, content usefulness, production complexity, and production quality. We additionally report the aggregate audio aesthetic score and text--audio similarity measured by MS-CLAP. As shown in Table~\ref{tab:audio_fidelity}, the distilled student maintains comparable audio quality and text--audio alignment to the teacher under 4-step inference, demonstrating that the proposed distillation framework effectively preserves audio generation capability while substantially reducing sampling steps.

\begin{table}[t]
\centering
\caption{
Audio quality and text--audio alignment evaluation between the original teacher and ours. The best result is \underline{underlined}.
}
\label{tab:audio_fidelity}
\papertablestyle
\resizebox{\columnwidth}{!}{%
\begin{tabular}{lcccccc}
\toprule
\textbf{Model}
& \textbf{Enjoy.} $\uparrow$
& \textbf{Useful.} $\uparrow$
& \textbf{Complex.} $\uparrow$
& \textbf{Prod. Qual.} $\uparrow$
& \textbf{Aes. Mean} $\uparrow$
& \textbf{MS-CLAP} $\uparrow$ \\
\midrule
Teacher (40 steps) & \underline{4.765} & 6.307 & \underline{3.275} & 6.577 & \underline{5.231} & \underline{0.353} \\
Ours (4 steps) & 4.612 & \underline{6.334} & 2.964 & \underline{6.584} & 5.115 & 0.347 \\
\bottomrule
\end{tabular}
}
\end{table}

\subsubsection{Diversity Evaluation}

We evaluate sampling diversity under fixed text conditions by generating 8 video-audio samples per prompt with different random seeds. Video and audio are encoded using a fixed pretrained ImageBind encoder, and pairwise embedding distances are computed only among samples from the same prompt. This yields $\binom{8}{2}=28$ pairs per prompt and 5,600 pairs per modality over 200 prompts. The final diversity score is the average within-prompt distance.


We report video and audio diversity using modality-specific ImageBind embeddings. For a compact video-audio summary, we additionally report their arithmetic mean, $D_{\mathrm{VA}}=(D_{\mathrm{video}}+D_{\mathrm{audio}})/2$. Since diversity alone does not measure generation quality, we report the Javis score in the same table to show the quality--diversity trade-off. As shown in Table~\ref{tab:diversity}, DMD-only achieves competitive quality but exhibits the lowest diversity, whereas sCM-only preserves the strongest diversity at the cost of substantially lower quality; by combining sCM and DMD, our staged approach achieves a more favorable quality--diversity balance.

\begin{table}[t]
\centering
\caption{
Diversity comparison under different distillation objectives. Best and second-best results are \textbf{bold} and \underline{underlined}.
}
\label{tab:diversity}
\papertablestyle
\begin{tabular*}{\linewidth}{@{\extracolsep{\fill}}lcccc@{}}
\toprule
\textbf{Model} &
\textbf{Javis} $\uparrow$ &
\textbf{Video Div.} $\uparrow$ &
\textbf{Audio Div.} $\uparrow$ &
\textbf{VA Avg. Div.} $\uparrow$ \\
\midrule
sCM-only & 0.1131 & \textbf{0.3026} & \textbf{0.5037} & \textbf{0.4032} \\
DMD-only & \underline{0.1812} & 0.1821 & 0.3561 & 0.2691 \\
\textbf{Ours (staged)} & \textbf{0.1963} & \underline{0.2322} & \underline{0.4197} & \underline{0.3259} \\
\bottomrule
\end{tabular*}
\end{table}

\begin{figure*}[t]
    \centering
    \includegraphics[width=\linewidth]{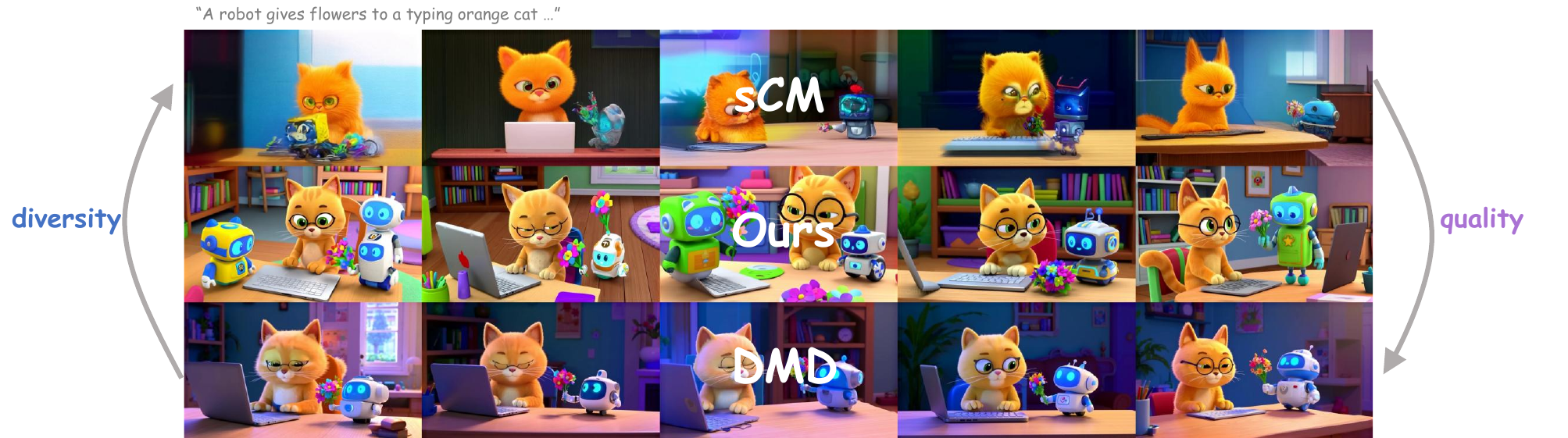}
    \caption{
    Quality--diversity comparison under matched prompts and random seeds. Rows show sCM-only, Ours (staged), and DMD-only, while columns show different random seeds. sCM-only provides broad variation but weaker prompt fidelity, DMD-only gives polished yet more repetitive compositions, and Ours maintains both coherent cat--robot interactions and meaningful cross-seed variation.
    }
    \label{fig:qualitative_results}
\end{figure*}

Figure~\ref{fig:qualitative_results} provides a controlled visual counterpart to the aggregate diversity scores. All three variants receive the same prompt and are sampled with the same set of random seeds. sCM-only changes subjects and layouts substantially, but its outputs less reliably preserve the requested cat--robot interaction. DMD-only produces polished frames but repeatedly converges to similar centered compositions. In contrast, Ours (staged) preserves the requested interaction while retaining visible cross-seed changes in character appearance, robot design, and scene layout. This qualitative comparison is consistent with the diversity trends observed in Table~\ref{tab:diversity}.

\subsubsection{High-Resolution Inference Acceleration}

\begin{figure}[t]
    \centering
    \includegraphics[width=\columnwidth]{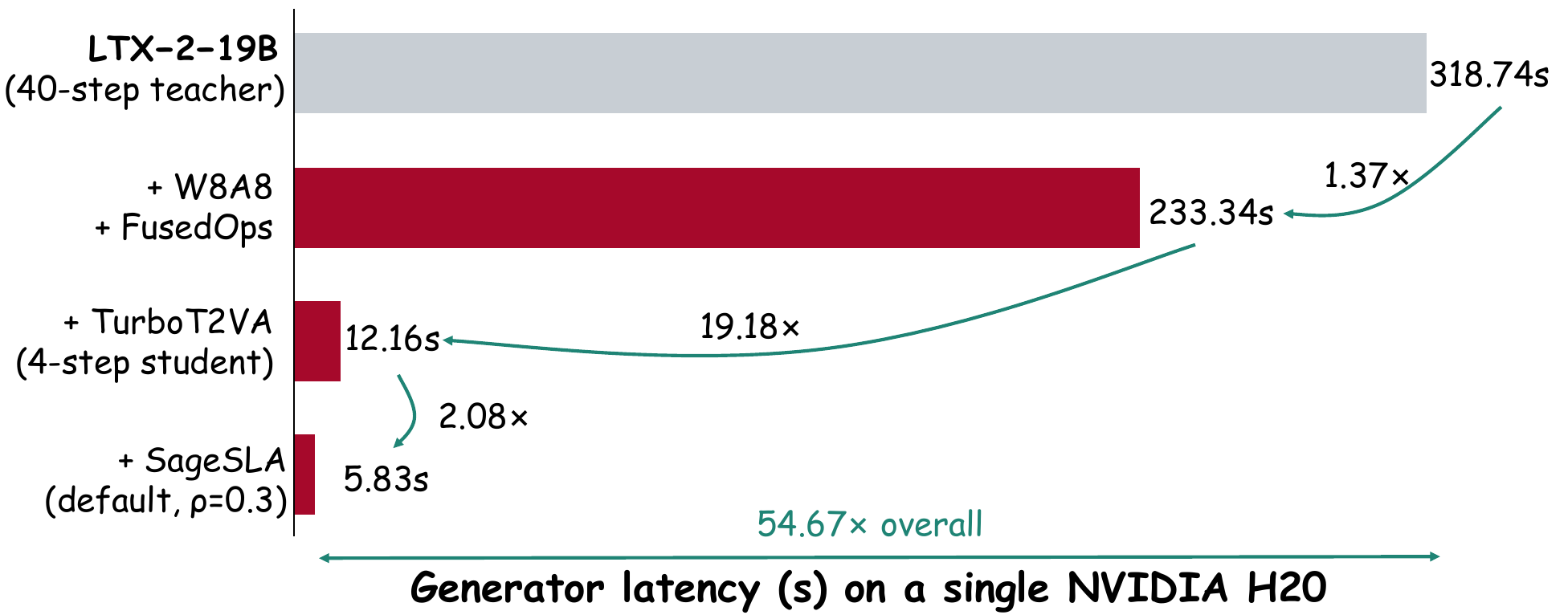}
    \caption{High-resolution inference acceleration, achieving a 54.67$\times$ generator-only speedup on one NVIDIA H20.}
    \label{fig:highres_acceleration}
\end{figure}

Figure~\ref{fig:highres_acceleration} evaluates the complete inference stack described in Section~\ref{sec:inference_acceleration} on one NVIDIA H20 at 1024$\times$1792, 121 frames, and batch size one. All stages are cumulative, and latency is measured for the generator only, excluding checkpoint loading, VAE decoding, muxing video and audio, and disk I/O. The dense 40-step teacher requires 318.74s per sample. Applying W8A8 linear operators and fused operations reduces this to 233.34s. Replacing the teacher trajectory with the four-step TurboT2VA student under the same operator stack yields 12.16s, corresponding to a 26.20$\times$ speedup over the dense teacher. Finally, modality-aware SageSLA with $\rho=0.3$ and padded-text compaction reduces latency to 5.83s, achieving a 54.67$\times$ overall generator speedup. For reference, the unoptimized four-step student takes 16.50s at this resolution, so the architecture-aware stack contributes an additional 2.83$\times$ acceleration beyond step reduction alone, demonstrating the complementary benefits of model-level distillation and per-step inference optimization. To isolate the contribution of individual inference components,
we further report the component-wise high-resolution latency in
Table~\ref{tab:highres_component_latency}. All student variants use
the same four-step inference setting, and only the enabled inference
components differ.

\begin{table*}[t]
\centering
\caption{
High-resolution generator latency breakdown at 1024$\times$1792 on one NVIDIA H20.
Best and second-best are \textbf{bold} and \underline{underlined}.
}
\label{tab:highres_component_latency}
\papertablestyle
\setlength{\tabcolsep}{4.5pt}
\resizebox{\textwidth}{!}{%
\begin{tabular}{@{}lcccccccccc@{}}
\toprule
\textbf{Metric}
& \textbf{Dense Teacher}
& \textbf{Dense Student}
& \textbf{Fused Only}
& \textbf{W8A8 Only}
& \textbf{W8A8 + Fused}
& \textbf{Sparse Only}
& \textbf{Full $\rho=0.5$}
& \textbf{Full $\rho=0.4$}
& \textbf{Full $\rho=0.3$}
& \textbf{Full $\rho=0.2$} \\
\midrule
Gen. (s) $\downarrow$
& 318.74
& 16.50
& 13.77
& 14.73
& 12.16
& 6.24
& 6.44
& 6.13
& \underline{5.83}
& \textbf{5.57} \\

Speedup $\uparrow$
& --
& 1.00$\times$
& 1.20$\times$
& 1.12$\times$
& 1.36$\times$
& 2.65$\times$
& 2.56$\times$
& 2.69$\times$
& \underline{2.83$\times$}
& \textbf{2.97$\times$} \\
\bottomrule
\end{tabular}}
\end{table*}

As shown in Table~\ref{tab:highres_component_latency},
fused operations and W8A8 independently reduce the dense four-step
generator latency from 16.50s to 13.77s and 14.73s, respectively,
while their combination further reduces it to 12.16s.
Sparse attention provides the largest single-component gain, and the
complete inference stack with $\rho=0.3$ reduces generator latency to
5.83s. These generator-only measurements isolate the contribution of individual inference components.

Beyond the component-wise latency analysis, we further characterize the
speed--quality trade-off by evaluating the complete inference stack under
different SageSLA retention ratios $\rho$. All configurations use the same four-step student, evaluation prompts, sampling schedule, and random seeds.

\begin{table}[t]
\centering
\caption{Speed--quality frontier of four-step TurboT2VA.
Best and second-best results are \textbf{bold} and \underline{underlined};
rankings use unrounded values.}
\label{tab:highres_speed_quality_frontier}
\papertablestyle
\setlength{\tabcolsep}{2.5pt}
\resizebox{\columnwidth}{!}{%
\begin{tabular}{@{}lcccccc@{}}
\toprule
\textbf{Config.} & \textbf{Gen. (s)}$\downarrow$
& \textbf{Javis}$\uparrow$ & \textbf{AVH}$\uparrow$
& \textbf{Desync}$\downarrow$ & \textbf{CAVP}$\uparrow$
& \textbf{IB-AV}$\uparrow$ \\
\midrule
Dense
& 16.50 & \textbf{0.1948} & \textbf{0.2330}
& 0.4275 & 0.7964 & 0.2370 \\
Full, $\rho{=}0.5$
& 6.44 & 0.1939 & 0.2287
& \textbf{0.3990} & 0.7974 & \underline{0.2375} \\
Full, $\rho{=}0.4$
& 6.13 & 0.1901 & 0.2247
& \underline{0.4080} & \underline{0.7974} & 0.2345 \\
Full, $\rho{=}0.3$
& \underline{5.83} & 0.1914 & 0.2269
& 0.4085 & \textbf{0.7985} & 0.2348 \\
Full, $\rho{=}0.2$
& \textbf{5.57} & \underline{0.1948} & \underline{0.2305}
& 0.4145 & 0.7971 & \textbf{0.2380} \\
\bottomrule
\end{tabular}}
\end{table}

As shown in Table~\ref{tab:highres_speed_quality_frontier}, reducing the retention ratio from $0.5$ to $0.2$ lowers generator latency from 6.44s to 5.57s, compared with 16.50s for dense four-step inference. Meanwhile, Javis, AVH, CAVP, and IB-AV remain close to the high-resolution dense reference across all retention ratios. We use $\rho=0.3$ as the default configuration because it provides a stronger balance between latency and cross-modal alignment, achieving the best CAVP score and lower desynchronization error than $\rho=0.2$. The more aggressive $\rho=0.2$ setting further reduces latency while maintaining comparable overall video-audio quality.

\begin{figure}[t]
    \centering
    \includegraphics[width=\columnwidth]{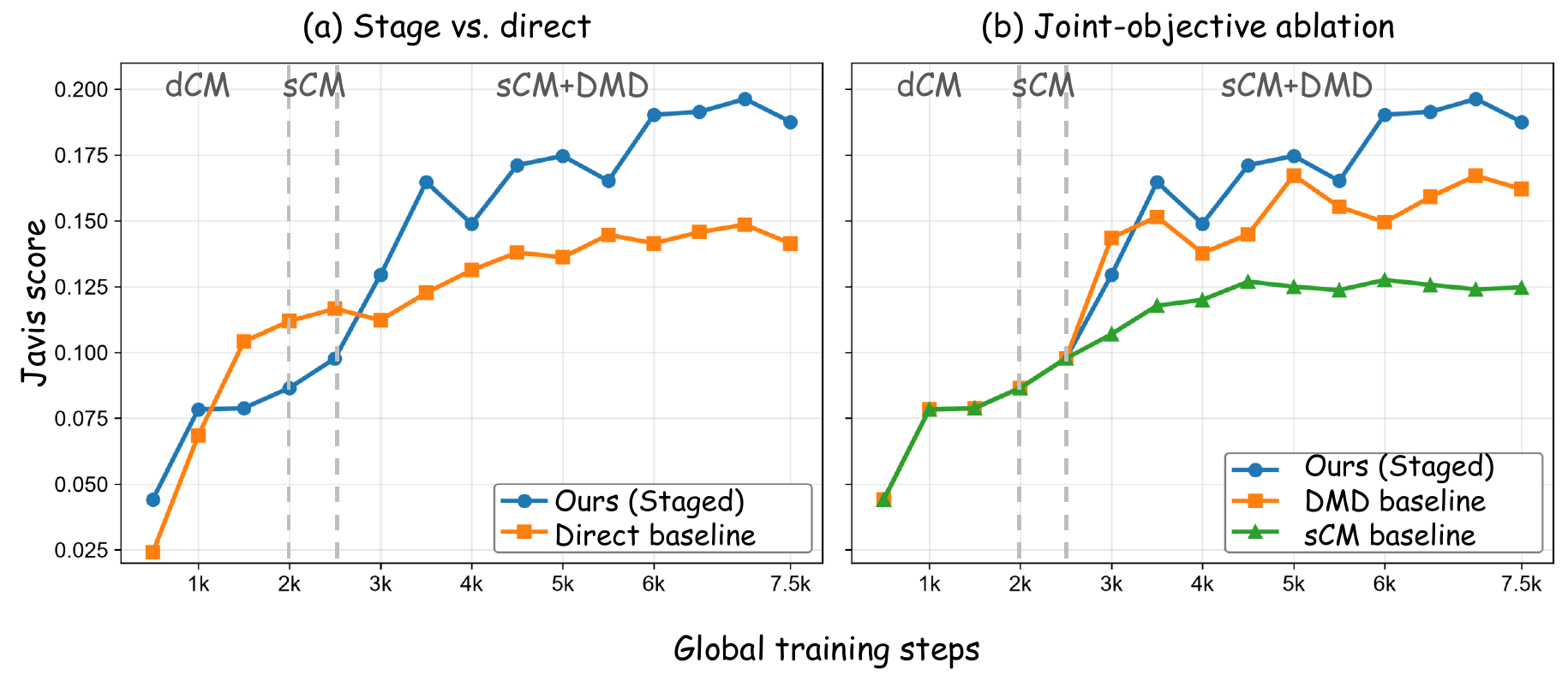}
    \caption{Training-curve ablations through 7,500 global steps. (a) Staged dCM→sCM→sCM+DMD training overtakes direct sCM+DMD optimization after entering the joint stage and maintains stronger late-training performance. (b) Starting from the same dCM+sCM prefix, staged sCM+DMD optimization achieves stronger late-training Javis scores than DMD refinement or sCM-only continuation. Dashed vertical lines denote the stage boundaries.    
    }
    \label{fig:combined_ablation}
\end{figure}
  
\subsection{Ablation Study}

We conduct ablation studies to validate the effectiveness of the progressive curriculum and the roles of sCM and DMD.

\subsubsection{Effect of Progressive Warm-up}

We first evaluate the effect of the warm-up stages. Directly optimizing sCM and DMD from the base initialization is trainable, but it does not yield the best quality--diversity trade-off. Without dCM warm-up and sCM refinement, the student receives trajectory-level and distribution-level supervision before acquiring a sufficiently structured few-step generation trajectory. In contrast, dCM warm-up provides a coarse denoising initialization, and sCM refinement further aligns the student with the continuous teacher trajectory before DMD is introduced. As shown in Figure~\ref{fig:combined_ablation}(a), the staged curriculum overtakes direct sCM+DMD optimization after the joint stage begins and retains a clear advantage through 7,500 global steps.

\subsubsection{Effect of Joint sCM+DMD Optimization}
Figure~\ref{fig:combined_ablation}(b) compares endpoint objectives after
the same dCM+sCM warm-up prefix. DMD refinement improves quickly after the
shared prefix but plateaus below the joint objective, whereas sCM-only
remains stable but substantially lower in Javis score. Staged sCM+DMD
optimization reaches the strongest late-training quality, peaking at
0.1963 at 7,000 steps and remaining ahead at the common 7,500-step endpoint.
Together with the diversity results in Table~\ref{tab:diversity}, these
curves show that joint consistency and distribution matching provide a
stronger quality--diversity balance without implying a fine-grained
objective-ratio sweep.

\begin{table*}[t]
\centering
\caption{
Sampling-step ablation for TurboT2VA. Best and second-best results are shown in \textbf{bold} and \underline{underlined}, respectively.
}
\label{tab:step_ablation_metrics}
\papertablestyle
\resizebox{\linewidth}{!}{
\begin{tabular}{c|cccccccccccc}
\toprule
\textbf{Steps} &
\textbf{Visual} $\uparrow$ &
\textbf{Motion} $\uparrow$ &
\textbf{Audio} $\uparrow$ &
\textbf{IB-TV} $\uparrow$ &
\textbf{IB-TA} $\uparrow$ &
\textbf{IB-AV} $\uparrow$ &
\textbf{CLIP} $\uparrow$ &
\textbf{CLAP} $\uparrow$ &
\textbf{CAVP} $\uparrow$ &
\textbf{AVH} $\uparrow$ &
\textbf{Javis} $\uparrow$ &
\textbf{Desync} $\downarrow$ \\
\midrule
1
& 2.0576
& 0.6680
& 4.0746
& \textbf{0.2259}
& 0.0919
& 0.0959
& 0.2915
& 0.2367
& 0.7955
& 0.0847
& 0.0663
& 0.6220 \\

2
& \underline{2.2328}
& \underline{0.7269}
& \underline{4.3416}
& \underline{0.2224}
& \underline{0.1377}
& \underline{0.1915}
& \textbf{0.2926}
& \underline{0.3198}
& \underline{0.7959}
& \underline{0.1845}
& \underline{0.1576}
& \textbf{0.3780} \\

4
& \textbf{2.3892}
& \textbf{0.8493}
& \textbf{4.4860}
& 0.2185
& \textbf{0.1447}
& \textbf{0.2290}
& \underline{0.2921}
& \textbf{0.3472}
& \textbf{0.8033}
& \textbf{0.2241}
& \textbf{0.1963}
& \underline{0.3880} \\
\bottomrule
\end{tabular}
}
\end{table*}

\subsection{Effect of Sampling Steps}

Finally, we evaluate TurboT2VA with different numbers of sampling steps. As shown in Table~\ref{tab:step_ablation_metrics}, generation quality generally improves as the number of sampling steps increases. Two-step inference achieves reasonable performance, while four-step inference provides further gains across most metrics.

\section{Conclusion}

In this paper, we present TurboT2VA, an efficient generation framework for large-scale text-to-video-audio synthesis. TurboT2VA extends score-regularized consistency distillation to a 19B-parameter joint video-audio model and addresses the modality imbalance and quality--diversity trade-off arising in few-step T2VA generation. The proposed modality-balanced objectives retain end-to-end optimization through the coupled video-audio backbone, while the progressive dCM$\rightarrow$sCM$\rightarrow$sCM+DMD curriculum gradually introduces trajectory-level consistency and distribution-matching supervision. The staged training strategy uses dCM to establish a stable coarse-denoising initialization, then switches to sCM for continuous teacher-trajectory learning before introducing distribution-level supervision.

Experiments on JavisBench, VBench, and modality-specific fidelity and diversity benchmarks demonstrate that the resulting four-step student achieves a favorable efficiency--quality trade-off. At the standard evaluation resolution of 512$\times$768, it reduces generator latency from 50.52s to 2.51s, corresponding to a 20.1$\times$ speedup, while retaining competitive visual quality, audio fidelity, cross-modal consistency, and sample diversity.

Beyond sampling-step reduction, TurboT2VA incorporates an architecture-aware inference stack consisting of guarded W8A8 linear operators, fused multimodal Transformer operations, padded-text compaction, and modality-aware sparse attention, while keeping cross-modal and text-conditioning attention paths dense. At 1024$\times$1792, the complete stack reduces generator latency from 318.74s to 5.83s on a single NVIDIA H20, achieving a 54.67$\times$ generator-only speedup. These results demonstrate the potential of combining few-step distillation with architecture-aware systems optimization for efficient deployment of large-scale joint video-audio generation models. Together, these model- and system-level improvements reduce both the number of Transformer evaluations and the computational cost of each evaluation in practical T2VA synthesis, while preserving the unified structure required for synchronized multimodal generation.

\bibliographystyle{IEEEtran}
\bibliography{references}

\end{document}